# Line and Word Matching in Old Documents

A. Marcolino*, V. Ramos**, M Ramalho*, J.Caldas Pinto*,
* IDMEC/IST - Technical University of Lisbon - Instituto Superior Técnico,
Av. Rovisco Pais, 1049-001 Lisboa, PORTUGAL
** CVRM/Centro de Geo-Sistemas - Instituto Superior Técnico
Av. Rovisco Pais, 1049-001 Lisboa, PORTUGAL

Abstract: This paper is concerned with the problem of establishing an index based on word matching. It is assumed that the book was digitised as better as possible and some pre-processing techniques were already applied as line orientation correction and some noise removal. However two main factor are responsible for being not possible to apply ordinary optical character recognition techniques (OCR): the presence of antique fonts and the degraded state of many characters due to unrecoverable original time degradation. In this paper we make a short introduction to word segmentation that involves finding the lines that characterise a word. After we discuss different approaches for word matching and how they can be combined to obtain an ordered list for candidate words for the matching. This discussion will be illustrated by examples.

## 1. Introduction

Among invaluable library documents are ancient books, which, beyond their rarity and value, represent the 'state of knowledge'. They constitute a heritage that should be preserved and used.

The fact that books must be handled with care to avoid damage does not allow that pages are flattened on the digitisation process, and thus geometrical distortions may appear. These should be corrected upon the digitisation process by the use of a suitable system, or a pre-processing analysis (e.g. [1]). In most books pages may be degraded due to humidity or other natural causes. This implies a pre-processing phase that includes geometric correction and/or cleaning or restoration. The book is then ready to be accessed. These books are used for rather specialised researchers. They look not only for indexing terms (key words), which sometimes are only meaningful for their particular research but also are interested in the presence or absence of specific graphic characteristics or other features. This work is concerned with this last problem. Two main factors makes it a challenging issue. The first one was already referred: the rarity of the books together with their state of conservation frequently originate bad quality digital versions. The second one in concerned with the printing process. Indeed old printed books are irregular in format and graphic quality. These factors together with the presence of typographical specific fonts make them unique exemplars. This address particular problems to recognition, which, cannot be supported by a dictionary or grammatical database.

This paper is divided in two main parts. In the first one we make a short survey of segmentation of text documents and present how we perform word segmentation and define line features. In the second part we discuss some techniques for word matching. It is expectable that matching techniques are unable to cope with all the problems that can affect digitised words, but we hope that all, or some of them, correctly combined will solve the problem with a higher degree of confidence. Examples are presented from images extracted from the first edition of *Lusíadas*, by *Luís Vaz de Camões* and *Les Observations de Plusieurs Singularitez,* by *Pierre Belon*, printed in 1445.

## 2. Line Segmentation

Line segmentation is an operation that belongs to the page segmentation class of problems. Although grey level images can be used, in general the segmentation is performed on binarised version of the images. This is almost compulsive when big amounts of pages are to be processed as in the present case (of the order of 500 pages). Assuming the text skew corrected projection profile cuts seems to be the most straightforward approach [10]. Indeed they reflect a global feature present in a significant part of the document and thus can be extensively and efficiently used.

In the current work projection profiles are used for segmentation of lines, words and characters. However, due to the nature of the texts, some nuances have to be taken which will be shortly described in the following sections.

**2.1 Base line and x-line estimation**

To carry out further word recognition, the position of the lines as illustrated in fig.1 should be known. These are, respectively, top line, x-line, base line and bottom line. However, it was sufficient to estimate two lines, as a heuristic rule can be applied. Indeed those lines follow approximately the following criteria:

$$top_line - x_line = x_line - baseline =$$

$$= baseline - bottom_line = k$$

Then to evaluate k we need only two lines. Early binding this type of restriction allowed us to speed up the process, at the cost of restringing the segmentation process to typographic structures similar to the one that are envisaged. This process could require further analysis, as the number of documents in the database is increased.

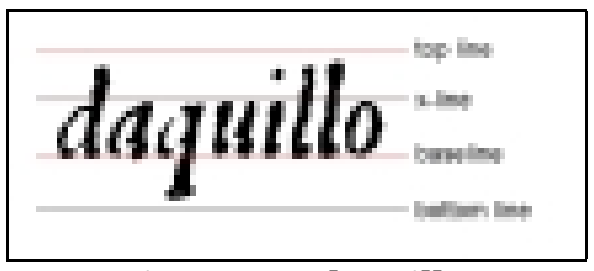

**Figure 1 - 'daquillo'**

Lines are extracted by thresholding the projection profile illustrated in fig. 2. However test samples do not present homogeneous characteristics in resolution or scale. This does not allow the threshold value to be fixed 'a priori' and automatic criteria must be fixed in order to simplify the process. If we look carefully to fig 2., we can see that the projection profile almost intuitively gives to us the lines of text that are formed by the image. If we imagine a simple vertical line to cross the entire image at the upper side of the projection profile, like the one in the figure, and then we walk through that hypothetical line from the top until the bottom, we will cross several times, two distinct areas. One area will be “empty” in the histogram, and the other area will be “full”. The transitions between the “empty and full areas” allow us to detect the baseline and the x-line of each line of text.

To trace the imaginary line we have used a statistical measure that we called the “mean value of pixels per line”:

$$\overline{mvpl} = \frac{\sum_{i=1}^{Height} x^i}{Height}$$

For the example of fig.2 a total of 8 lines were extracted.

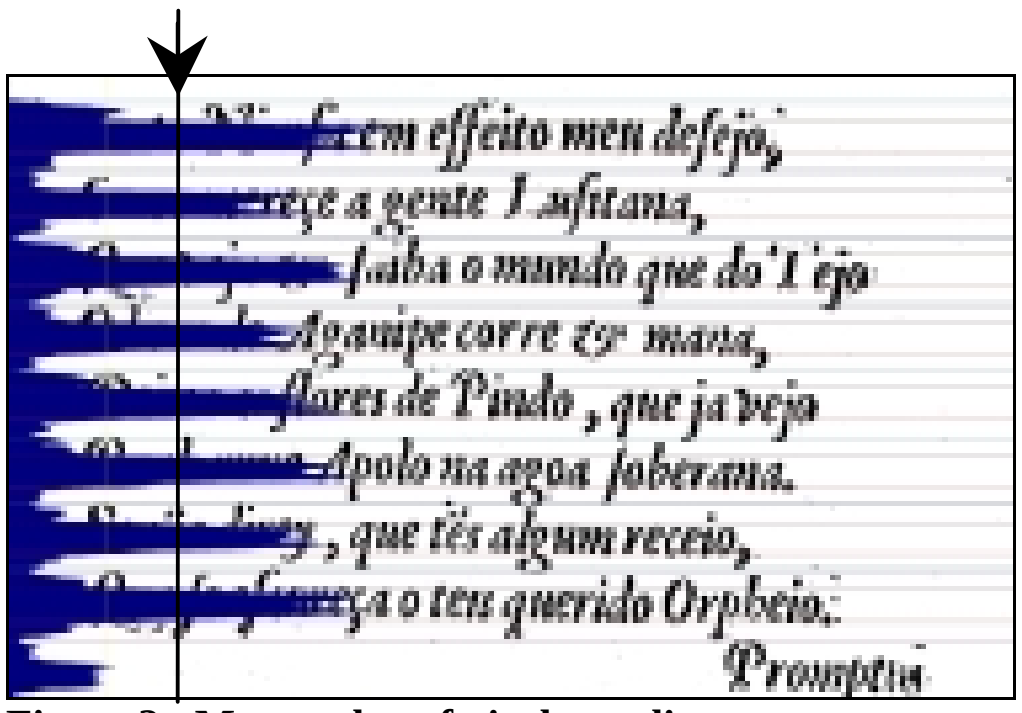

**Figure 2 - Mean value of pixels per line**

That approach still leaves some cases unattended. Unexpected shorter lines, as that shown in Figure 3, are not detected. However, this error can be easily corrected if we reapply the same algorithm to the portions of the image that didn’t contribute with any line. Thus shorter lines are detected.

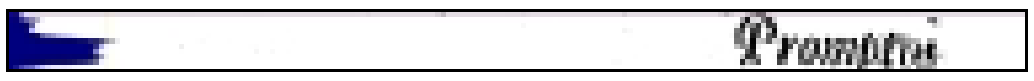

**Figure 3 - Shorter line**

A second kind of problem arises when we walk through the projection profile, along the mean value. A single line may be cut in two or more 'lines'. This problem is corrected through a voting procedure. Each line votes with the corresponding height to choose a representative. Majority elects the representative lines and those that differ significantly (>50%) from the representative are discarded.

## 3. Word Segmentation

This process has some steps that are very similar to the problem of line segmentation. The process starts by computing the projection profile of the image, iterating across vertical lines. With this projection profile, the histogram of the space length is build. Two distinct areas are clearly present. The space between characters corresponds to the smaller bin values. The spaces between words correspond to higher bin values. Thus it is necessary to find out an objective criterion for automatic thresholding.

In our approach this is achieved trough the so-called histogram of lengths. We assume that two characters are separated by a very small amount of pixels. On the other hand, we assume that two words are separated by a significant amount of pixels. With the projection profile we can construct a histogram of the lengths of white spaces that are in the line. An example can be seen on Fig. 4.

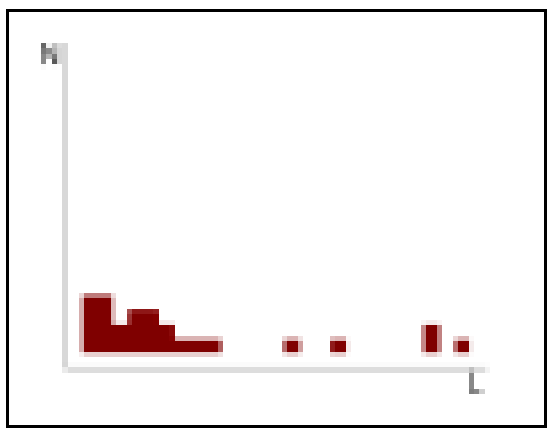

**Figure 4 - Length Histogram**

We can interpret this histogram to draw some conclusions. The left side of the histogram is where are represented the spaces between the characters, that correspond to small widths. The right side of the histogram is where are represented the spaces between the words, that correspond to bigger widths. After a closer analysis we conclude that by simply pushing the mean-value to the right side of the histogram until we find a null value an acceptable solution is attained.

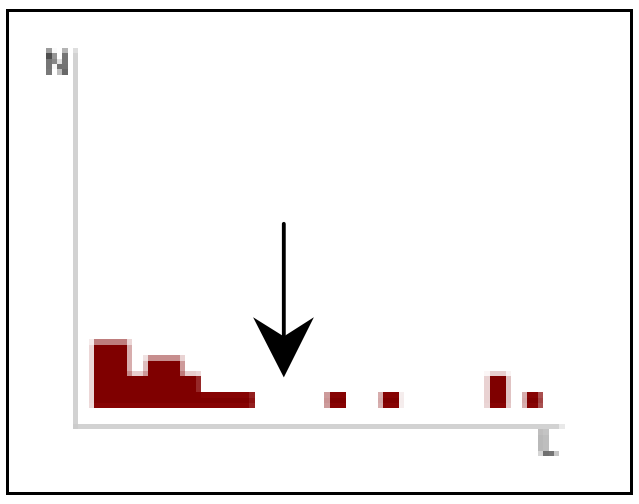

**Fig. 5 - Threshold**

The final result can be seen in Fig. 6. A total of 6 words were extracted.

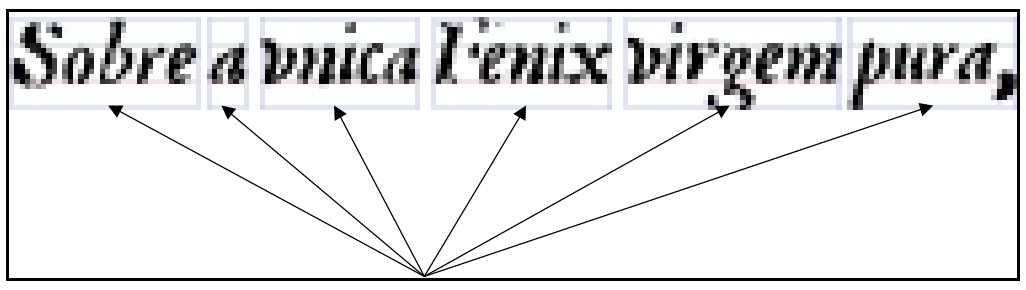

**Figure 6– *'Sobre a unica Fenix virgem pura,'***

# 4. Word Recognition

### 4.1. Introduction

Several solutions have been proposed with the specific job of doing word recognition, which is related in some ways with character recognition. However, the difficulty of this type of task is enormous, and increases at the same time, that the degree of degradation of the text, of the pictures scanned, increases.
It was not the aim of this work to convert the text of the scanned pictures to characters codes, e.g. ascii. The analysed documents contain symbols that are not used anymore as the one in figure 6.
One of the techniques to carry out the recognition consists of extracting a set of features from the pictures, and then use that information to obtain the classification desired. Several features and methods. have been proposed (e.g. strokes, contour analysis) [2][3][5][7][8]. However in general, the quality of the results quickly deteriorate as the degradation of the pictures increases.

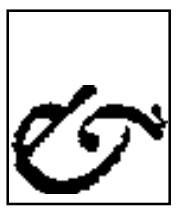

**Figure 6 – Old fonts**

In this paper we are going to present some steps already given to tackle this problem. We will present different techniques for word matching and perform some comparisons. We believe that only a combination of features will tend to solve the problem but this will be the next step of this work.
At this point we start with the information of the bounding boxes that limit the words. The task involves selecting a picture of a word and presents all the words that look alike.

### 4.2. World length threshold

This is the simpler feature. It seems obvious that a matching candidate word should have approximately the same length as the word to match. To images scanned at 300 dpi it was find out that a tolerance of ±10 pixels it's sufficient to cover all correct matches in the same font, including, obviously a large number of incorrect candidates. The number of indexing terms that can be lost due to different text characteristics (different font size or character set) didn't appear to be significant, and thus refinement's were left for a later project development stage.

### 4.3. Correlation between words

Another process to be taken into account is correlation. Indeed, as there are no severe problems of distortions or rotations between words (those have been eliminated in the pre-processing stage) it allows a further refinement of the candidates that has been precious for project feasibility. Correlation is calculated as, for two word enveloping blocks $I_1$ and $I_2$, as

$$SSD = \frac{\sum (I_1 - I_2)^2}{\sqrt{\sum I_1^2 \sum I_2^2}}$$

Results have shown that correlation is a good measure and should have a significant degree of believe. Another advantage of this method is the execution time, which is very fast. This is of great importance when dealing with books consisting of more than four hundred pages.
Results for some words are shown in fig.7. In certain cases the discrimination almost suggest that we do not need any other measure. However, as it can be seen in the same figure, and in other examples, the existence of false positives, will force us to look for another descriptors.

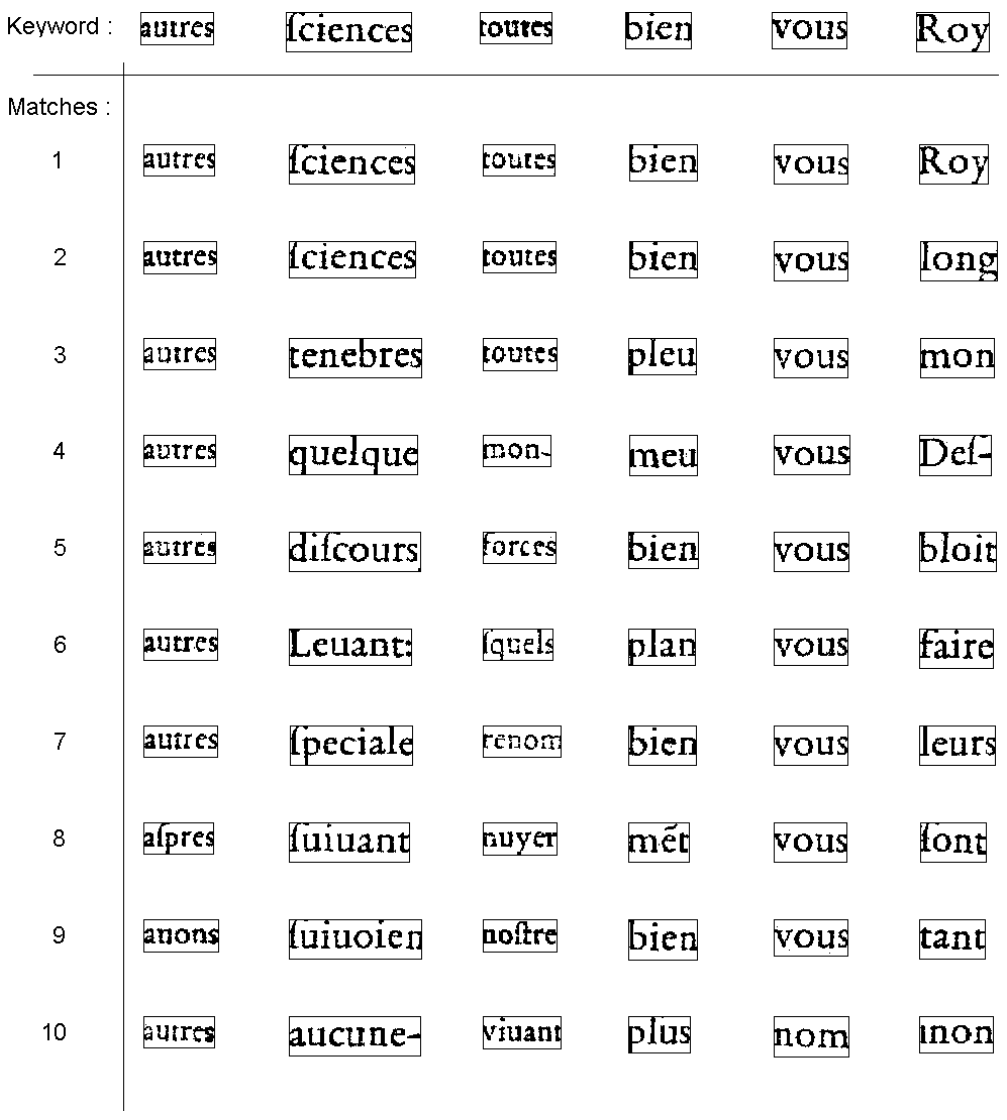

**Fig. 7. Matching results using correlation results are then ordered by increasing order of similarity.**

### 4.4. Character Shape Code

Another fast method that at the first sight could be used to help the discrimination of candidates is the so-called character shape code from Splitz [9].
In this method each character image is divided in three zones (fig. 8) and classified accordingly.

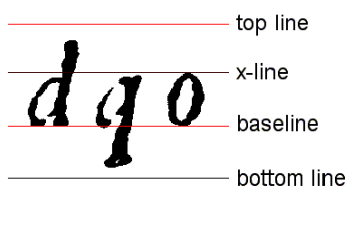

**Figure 8 Zones in a character image**

Characters that have pixels between the top line and the x-line are considered ascenders. Characters that have pixels between the baseline and the bottom line are considered descenders. The second character of the previous figure is an example. Last, characters that have all the pixels between the x-line and the baseline are neither ascenders nor descenders.
A signature based on ascenders and on descenders can this way be extracted and possibly used to select or discard further words. Unfortunately broken and touching characters occurs sparingly. The codification in ascenders and descenders in these cases may not be applied character by character. This method was then modified to extract the information about ascenders and descenders at fixed points in a word's image, bypassing the character segmentation phase. The word image can be divided in sectors with a specified width. A width of 15 pixels, which approximately corresponds to one character, is used. This is illustrated on fig. 9. For this image the codified string would be AAxxxxx. An 'A' designates a sector with an ascender. A 'x' designates a sector that is neither an ascender nor a descender. A 'D' designates a sector with a descender.

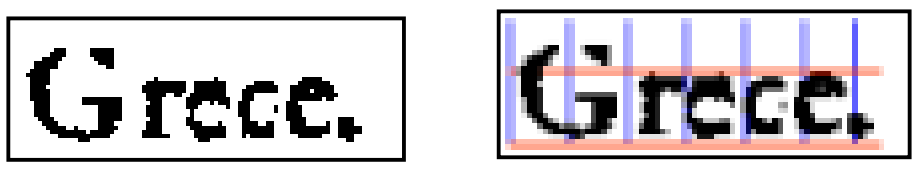

**Figure 9 Shape code string evaluation**

A simpler measure is the number of mismatches between the shape codes of the words to be compared.
As already stated the pictures of the words have a poor binarization that causes problems in the extraction of the signatures. With incorrect signatures taken from the words this method is prone to errors that were not desired. Our conclusion is that this method may help to detect some intruders in the list obtained using correlation. This will be tested in section 4.

### 4.5. Ulam's Distance

*Bhat* [7] has presented an evolutionary measure for image matching that is based on the *Ulam's* distance - a well know ordinal measure from molecular biology, based on an evolutionary distance metric that is used for comparing real DNA strings. Given two strings, the *Ulam's* distance is the smallest number of mutations, insertions, and deletions that can be made within both strings such that the resulting substrings are identical. Then, *Bhat* reinterprets the *Ulam's* distance with respect to permutations that represent windows intensities expressed on an ordinal scale. The motivation for him to use this measure is twofold: it not only gives a robust measure of correlation between windows but also helps in identifying pixels that contribute to the agreement (or disagreement) between the windows.
Given a set of two window ($I_1$, $I_2$) intensity values given by $(I^i_1, I^i_2)^n_{i=1}$, let $\pi^i_1$ be the rank of $I^i_1$ among the $I_1$ data, and $\pi^i_2$ be the rank of $I^i_2$ among the $I_2$ data. Consider for instance the following example with two 3 X 3 windows, being $n = 9$. Then $(I^i_1)^n_{i=1}$= :

| | | |
|---|---|---|
| 10 | 30 | 70 |
| 20 | 50 | 80 |
| 40 | 60 | 100 |

i.e. $(I^i_1)_{i=1}$=10, $(I^i_1)_{i=2}$=30, …, $(I^i_1)_{i=9}$=100. For $(I^i_2)^n_{i=1}$ we have:

| | | |
|---|---|---|
| 10 | 30 | 70 |
| 20 | 50 | 80 |
| 40 | 60 | 15 |

Thus we have for $\pi^i_1$:

| | | |
|---|---|---|
| 1 | 3 | 7 |
| 2 | 5 | 8 |
| 4 | 6 | 9 |

and for $\pi^i_2$ the following rank matrix:

| | | |
|---|---|---|
| 1 | 4 | 8 |
| 3 | 6 | 9 |
| 5 | 7 | 2 |

we then define a composition permutation $s^i$ as:

$$\left(s^i\right)_{i=1}^9 = \pi_2^{i:\pi_1^i=1}, \pi_2^{i:\pi_1^i=2}, \ldots, \pi_2^{i:\pi_1^i=9}$$

i.e., informally $s^i$ is the rank of the pixel in $I_2$ that corresponds to the pixel with rank $i$ in $I_1$. That is:

$$\left(s^i\right)_{i=1}^9 = (1,3,4,5,6,7,8,9,2)$$

Under perfect positive correlation between two windows, $s^i$ should be identical to the *identity permutation* given by:

$$\left(u^i\right)_{i=1}^9 = (1,2,3,4,5,6,7,8,9)$$

Under perfect negative correlation, i.e. the sequences completely disagree, $s^i$ must be identical to the *reverse identity permutation* given by:

$$\left(r^i\right)_{i=1}^9 = (9,8,7,6,5,4,3,2,1)$$

The *Ulam's* distance $\delta$, between the permutations $s^i$ and $u^i$ is defined as the minimum number of elements that must be removed from each permutation such the resulting subsequences agree perfectly. This is also equal to $n$ minus the length of the longest increasing subsequence in $s^i$. For obtaining the distance between $s^i$ and $r^i$, we construct $s^{*i}$ as:

$$s^{*i} = s^{n-i+1}, i = 1,\ldots,n$$

$$\left(s^{*i}\right)_{i=1}^9 = (2,9,8,7,6,5,4,3,1)$$

The *Ulam's* distance between $s^i$ and $r^i$ is equal to that between $s^{*i}$ and $u^i$. For our example we then have the longest common increasing subsequence between $s^i$ and $u^i$ as (1,3,4,5,6,7,8,9) and between $s^{*i}$ and $u^i$ as (2,9). Then we have the *Ulam's* distance between $s^i$ and $u^i$ , $\delta_1 = \delta( s^i, u^i) = 9 - 8 = 1$. Similarly, the

$$\tau_u = 1 - \frac{2\delta_1}{n-1} = 0.75$$

*Ulam's* distance between $s^{*i}$ and $u^i$ , $\delta_2 = \delta( s^{*i}, u^i) = 9 - 2 = 7$. Both $\delta_1$ and $\delta_2$ can take values in the range $[0,n\text{-}1]$. Then two measures of correlation $\tau_u$ and $\tau_r$ can be defined as:

$$\tau_r = 1 - \frac{2\delta_2}{n-1} = -0.75$$

$$\tau = \frac{\tau_u - \tau_r}{2} = 0.75$$

Both are distributed in the range [-1,1]. To obtain a measure that is symmetrically distributed around zero, we can define an average quantity $\tau$ as:

The previous approach, even if robust, has some drawbacks. The problem is to define the appropriate ranking matrixes for tied ranks, which although can be possible within grey level intensities, is more probable to happen in the case of binary images. If this is the case, we must then use another strategy to code the ranking matrixes. One possible strategy is to compute the $(\pi^i_1, \pi^i_2)^n_{i=1}$ rankings assuming that the importance for each tied group of cells are redefined by the following order**:**

| | | |
|---|---|---|
| 1 | 2 | 3 |
| 4 | 5 | 6 |
| 7 | 8 | 9 |

To deal the same problem, although not mentioned in [7], *Bhat* and *Nayar* [1] ordered tied pixels in raster scan fashion, which is similar to the present strategy. It makes sense as it emphasises positive correlation between corresponding windows. However, it should be noticed that one of the drawbacks of ordinal measures, in general, is that we lose information by going to an ordinal scale. For grey-level images, by viewing intensity as an ordinal variable we lose the grey-level ratio information between pixels. In the case of binary images, the present approach seems to be promising because we are not losing any information going into the ordinal scale.

### 3.5 Character segmentation

The techniques developed to do word segmentation can be applied to segment the characters in a given word. The number of obtained characters (even if incorrect due to typographical specificities) may be another feature that can contribute to refine the algorithm.

## 5. Results

Algorithms presented on this paper will now be tested in order to check their possible contribution to a decision rule that can result from a combination of these descriptors. Results are presented along with tables corresponding to examples analysed. In the first column words are aligned according with the correlation results. In the remaining ones how the orders descriptors evaluate the distance of the corresponding word to the key one. This will give us some insight about how these features may help to correct results obtained by the correlation technique.

**Table I: Results for the word ”autres”**

| Keyword : | autres |
|---|---|
| Matches : | |
| 1 | autres |
| 2 | autres |
| 3 | autres |
| 4 | autres |
| 5 | autres |
| 6 | autres |
| 7 | autres |
| 8 | afpres |
| 9 | auons |
| 10 | autres |

**Table II: Results for the word ”bien”**

| Keyword : | bien |
|---|---|
| Matches : | |
| 1 | bien |
| 2 | bien |
| 3 | pleu |
| 4 | meu |
| 5 | bien |
| 6 | plan |
| 7 | bien |
| 8 | mét |
| 9 | bien |
| 10 | plus |

**Table II: Results for the word ”defcription”**

| Keyword : | |
|---|---|
| Matches : | |
| 1 | defcription |
| 2 | defcription |
| 3 | demeurera |
| 4 | s'efforçants |

## 6. Conclusions

The task of doing word recognition without the aid of the ‘semantic level’ is problematic. Humans and computers don’t see symbols in the same way. Yet all the methods try to imitate more or less an imaginary process of how we recognise symbols. However we know that our brain performs a fusion of all the information captured by our different senses. The same will have to be done here. Fuzzy partnership relations combined with neural networks decision rules, together with further developed descriptors are expected to give improved results in our proposed task of word indexing in ancient books. This will be our goal as future work.

## 7. Acknowledgments

This project is supported by the European Project LB5608/A - DEBORA

## 8. References

[1] Ramalho M, Caldas Pinto J, Cabral R; Machado O; *Antique Book Page Geometrical Restoration,* Recpad 2000, Porto, May 2000, Portugal.

[2] Sung-Hyuk Cha et al., *Approximate Stroke Sequence String Matching Algorithm for Character Recognition and Analysis*, in Proc. 5th Int. Conf. On Document Analysis and Recognition, ICDAR'99, Bangalore, India, pp.53-56.

[3] David Doerman, *The Detection of Duplicates in Document Image Databases*, in Proc. 4th Int. Conf. On Document Analysis and Recognition, ICDAR'97.

[4] C.H. Chen and J.L. DeCurtins, *Word Recognition in a Segmentation-free Approach to OCR*, in Proc. 2 th Int. Conf. On Document Analysis and Recognition, ICDAR'91.

[5] Dinkar Bhat, *An Evolutionary Measure for Image Matching*, in Proc. 14th Int. Conf. On Pattern Recognition, ICPR'98, Brisbane, Australia 1998, Vol I, pp.850-852.

[6] Vitorino Ramos, An Evolutionary Measure for Image Matching – Extensions to Binary Image Matching, Internal Technical Report CVRM/IST 2000.

[7] Didier Guillevic and Ching Y. Suen. HMM Word Recognition Engine, in Proc. 4th Int. Conf. on Document Analysis and Recognition, ICDAR'97, Ulm, Germany, Vol. 2, pp. 544-547, August 1997.

[8] Didier Guillevic, *Unconstrained Handwriting Recognition Applied to the Processing of Bank Cheques*, Doctoral thesis, Computer Science Department, Concordia University, Montreal, September 1995.

[9] AL Spitz, Shape-based word Recognition, International Journal on Document Analysis and Recognition, vol 1, nº 4 pp 178-190, 1999

[10] Tang YY, Lee SW, Suen CY, *Automatic Document Processing: A survey*; Pattern Recognition, 29(12), pp 1931-1952, 1996